\documentclass[a4paper]{esannV2}
\usepackage{graphicx}
\usepackage[utf8]{inputenc}
\usepackage{amssymb,amsmath,array}
\usepackage{booktabs}
\usepackage{multirow}
\usepackage{xcolor}
\usepackage[numbers,sort&compress]{natbib}
\usepackage{hyperref}
\hypersetup{hidelinks}

\newcommand{\vobs}{v_{\mathrm{obs}}}
\newcommand{\vinh}{v_{\mathrm{inh}}}
\newcommand{\vtot}{v_{\mathrm{tot}}}
\newcommand{\E}{\mathbb{E}}
\newcommand{\Var}{\operatorname{Var}}

\newcommand{\Figref}[1]{Figure~\ref{#1}}

\newcommand{\tabref}[1]{Table~\ref{#1}}

\usepackage[acronym,nomain,nopostdot]{glossaries}
\makenoidxglossaries
\glsdisablehyper
\newacronym{rl}{RL}{reinforcement learning}
\newacronym{pomdp}{POMDP}{partially observable Markov decision process}
\newacronym{td}{TD}{temporal-difference}
\newacronym{dqn}{DQN}{deep Q-network}

\pdfpagewidth=\paperwidth \pdfpageheight=\paperheight %
\begin{document}
\setlength{\bibsep}{0pt}%
\title{EXPOSURE: Explaining Sources of Uncertainty in Reinforcement Learning\thanks{Submitted to ESANN 2027, 35th European Symposium on Artificial Neural Networks, Computational Intelligence and Machine Learning, Bruges, Belgium, 21--23 April 2027.}}

\author{Kiana Farhadyar$^{1}$, Yannick Vogt$^{1,2}$, Philipp Laur$^1$, Ay\c{c}a \c{C}elik$^1$, \\
      Iman Nematollahi$^1$, Gabriel Kalweit$^{3,1}$ and Maria Kalweit$^{1,2,3}$
      \thanks{This work is partly funded by BrainLinks-BrainTools (Federal Ministry of Economics, Science and
            Arts of Baden-W\"urttemberg, sustainability program of the Excellence Initiative II) and supported by
            CRIION gGmbH (Collaborative Research Institute Intelligent Oncology).}
      \vspace{.3cm}\\
      1- Department of Computer Science, University of Freiburg, Freiburg, Germany
      \vspace{.1cm}\\
      2- IMBIT//BrainLinks-BrainTools, Freiburg, Germany
      \vspace{.1cm}\\
      3- Collaborative Research Institute Intelligent Oncology, Freiburg, Germany
}

\maketitle

\begin{abstract}
 Shapley values in reinforcement learning are used to explain which observation features drive an agent's behavior and expected return, but not its certainty. Explaining certainty requires understanding how observation features influence different sources of return uncertainty, and  Shapley values need the agent's prediction for every subset of visible features. We propose EXPOSURE, which uses one mask-aware belief network for both action selection and Shapley explanations. It explains the variance of the return, split into an observational part due to missing features and an inherent part due to the environment's own randomness, and the entropy of its belief over the masked features. In tic-tac-toe and Minesweeper, where exact Shapley values are computable, EXPOSURE gives more accurate explanations of expected return and its uncertainty, while uniquely explaining belief uncertainty.
\end{abstract}

\section{Introduction}
\label{sec:intro}

In a given state, two actions may have the same expected return, yet one may be risky while the other guarantees success. A \gls{rl} agent that explains only its expected return cannot tell them apart. In safety-critical applications such as healthcare, the agent's uncertainty and its sources matter. Assessing an agent's reliability requires understanding whether its uncertainty about expected return arises from observation features it does not see or from randomness inherent to the environment. It is also essential to know whether the observation features behind the agent's certainty are plausible or just a shortcut. Shapley values~\citep{shapley1953} are commonly used to explain supervised models~\citep{lundberg2017shap}, including their predictive uncertainty~\citep{watson2023infoshap}. In \gls{rl}, they explain an agent's decisions and outcomes~\citep{beechey2023sverl,beechey2025framework}, but the sources of its uncertainty remain largely unexplored.
\looseness=-1 

Every Shapley explanation requires a value for each coalition, i.e., each subset of visible features, and these values form a cooperative game. For each coalition, the uncertainty splits into an observational part, caused by the masked features, and an inherent part, which remains even when the agent sees all features, for example because the opponent plays randomly. Independent of the expected return, the agent also forms a belief, i.e., a probability distribution over the values of the masked features, and this belief can itself be uncertain. From experience, the agent learns which completions of a coalition are likely, and the entropy of its belief reflects how many remain plausible. Attributing each of these uncertainty targets answers its own question, namely \emph{which observation features make the agent less uncertain of the action's value?} (observational), \emph{which ones reveal that less uncertainty remains in the rest of the game?} (inherent) and \emph{which ones narrow the agent's belief about the masked features?} (belief).
\looseness=-1 

\begin{figure}[tb]
\centering
\includegraphics[width=\textwidth]{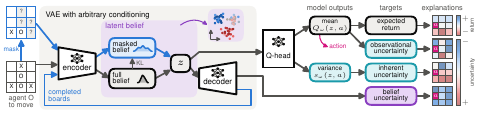}
\caption{Overview of EXPOSURE. An encoder maps a board with any subset of features masked to a latent belief, trained to match the belief from the full board. Blue arrows trace the masked board's path, while the agent acts on the belief from the full board. From a belief, a Q-head predicts the mean and variance of the return, and a decoder completes the masked features. The maps show each feature's Shapley value for the four explanation targets on the input board.}
\label{fig:arch}
\end{figure}

The common approach to obtain the coalition values is to train a separate surrogate network to predict the agent's values from masked inputs~\citep{covert2021removing}, so the explanations describe the surrogate rather than the agent. We instead propose EXPOSURE, an agent built on a mask-aware belief network trained under random masks (\Figref{fig:arch}). The agent acts on its belief from the full observation, and the explanations use its belief from each coalition, so they come from the agent itself. We evaluate it in tic-tac-toe and Minesweeper $4\times4$, where exact Shapley values can be computed. We compare with the surrogate approach and with a mask-aware agent without a latent belief that maps masked inputs directly to expected return, which isolates the benefit of the belief. In both environments, EXPOSURE gives the most accurate explanations of every target it shares with the baselines, and it alone explains the belief uncertainty. In partially observable Minesweeper, it achieves a higher win rate than baselines, while being competitive in tic-tac-toe.

Our contributions are (i) Shapley explanations of the observational, inherent and belief uncertainty of an \gls{rl} agent, (ii) EXPOSURE, a single belief network for both acting and explaining, and (iii) an evaluation against exact explanations in tic-tac-toe and Minesweeper.

\section{One mask-aware belief network}
\label{sec:model}

EXPOSURE is one mask-aware network with three parts: an encoder that forms the belief, a Q-head that predicts the return, and a decoder that completes the masked features. Let an observation $o$ have features $N = \{1,\dots,n\}$ and let $a$ be a legal action. For each coalition $C\subseteq N$, the encoder $e_\psi(z \mid o_C, C)$ maps the masked observation $o_C$ and $C$ to a mixture of $M$ diagonal Gaussians over a latent variable $z \in \mathbb{R}^d$, with weights $w_m$, means $\mu_m$ and scales $\sigma_m$. With nothing masked ($C = N$), it gives the full-observation belief. The Q-head estimates the expected return $Q_\omega(z,a)$ and return variance $s_\omega(z,a) \ge 0$, and the decoder $p_\theta(o_{N\setminus C} \mid z, o_C)$ reconstructs the masked features. During training, we draw replay transitions $(o, a, r, o')$ with next observation $o'$ and a random coalition $C$ for each, and minimize $\mathcal{L}_{\mathrm{belief}} + \lambda\,\mathcal{L}_{\mathrm{TD}}$ with weight $\lambda$, where
\begin{align}
\mathcal{L}_{\mathrm{belief}} &= \E_C\Big[\mathrm{KL}\big(e_\psi(\cdot \mid o_N)\,\|\,e_\psi(\cdot \mid o_C)\big) - \E_{z}\log p_\theta(o_{N\setminus C} \mid z, o_C)\Big], \label{eq:belief}\\
\mathcal{L}_{\mathrm{TD}} &= \E_{z}\Big[\big(Q_\omega(z,a) - y\big)^2 + \big(s_\omega(z,a) - \varsigma\big)^2\Big]. \label{eq:td}
\end{align}
The belief loss follows the variational autoencoder with arbitrary conditioning~\citep{ivanov2019vaeac, olsen2022vaeac}, but with one encoder for both posterior and conditional prior.
In the \gls{td} loss, $y=r+(1-t)\max_{a'}\bar Q(o',a')$ is the target for the undiscounted return and $\varsigma=(y-\bar Q(o,a))^2+(1-t)\bar s(o',a^*)$ the sampled return-variance target~\citep{sherstan2018variance}. Here $t$ indicates termination, $a^*$ is the greedy next action, and $\bar Q,\bar s$ are full-observation averages of the target encoder and head. Both losses use one reparameterized sample per mixture component of $e_\psi(\cdot\mid o_N)$, weighted by $w_m$, and the KL is estimated with these samples and exact mixture densities. The agent acts greedily on the full-observation mean $Q(o,a)=\E_z[Q_\omega(z,a)]$.
\looseness=-1 

\section{Certainty games}
\label{sec:games}
We now define the games, i.e., the coalition values that the Shapley values attribute, and how EXPOSURE estimates them. Let $G$ be the return after taking action $a$ and following a policy $\pi$, with $Q^\pi(O,a) = \E_\pi[G \mid O, a]$, and let $O$ follow the occupancy of $\pi$, i.e., how often each observation occurs when acting under $\pi$. For a coalition $C$, the law of total variance gives
\begin{equation}
\underbrace{\Var_\pi(G \mid o_C, a)}_{\vtot(C)} = \underbrace{\Var\big(Q^\pi(O,a) \mid o_C\big)}_{\vobs(C)} + \underbrace{\E\big[\Var_\pi(G \mid O, a) \mid o_C\big]}_{\vinh(C)}.
\label{eq:split}
\end{equation}

Observational uncertainty $\vobs$ is caused by hiding $N \setminus C$ and is zero at $C = N$. Inherent uncertainty $\vinh(N)$ persists even under full observation, so unmasking features cannot reduce it, but reveals how much of it remains, e.g., little on a nearly decided board. In common terms, $\vinh$ is aleatoric given the full observation, while $\vobs$ is not epistemic in the sense of model uncertainty, since a perfectly trained agent still has it when features are masked. Such an agent would estimate $\vtot(C)$ without separating its two parts, which EXPOSURE does through its belief. For a trained agent, its full-observation predictions $Q$ and $s$ replace $Q^\pi$ and $\Var_\pi(G \mid O, a)$. Besides these two uncertainties, we explain the expected return $\E[Q^\pi(O,a) \mid o_C]$, as in~\citep{beechey2023sverl,beechey2025framework}. Finally, the belief game measures how uncertain the agent is about the masked features. Its value is the entropy given the visible ones under the occupancy of $\pi$, $H(C) = \mathbb{H}[O_{N\setminus C} \mid o_C]$, which is zero at $C = N$. Unlike $\vobs$, $H$ also counts completions with the same return, so a feature can lower $H$ without changing $\vobs$.

The Shapley value $\phi_i$ of feature $i$ in a game $u$ is its marginal contribution $u(C \cup \{i\}) - u(C)$, averaged over all orders in which the features are revealed~\citep{shapley1953}, and a superscript names the game, e.g., $\phi^{\mathrm{obs}}$. We explain uncertainty reduction, $u(C) = v(\emptyset) - v(C)$, so $\phi_i > 0$ means seeing feature $i$ makes the agent more certain. By linearity, $\phi^{\mathrm{tot}} = \phi^{\mathrm{obs}} + \phi^{\mathrm{inh}}$, and efficiency gives $\sum_i \phi^{\mathrm{obs}}_i = \vobs(\emptyset)$. \looseness=-1 

EXPOSURE estimates each game from its masked belief $z\sim e_\psi(\cdot\mid o_C)$, with $\hat Q(C)=\E_z[Q_\omega]$ and $\hat v_{\mathrm{inh}}(C)=\E_z[s_\omega]$. For $\hat v_{\mathrm{obs}}$, we complete the masked features with decoder samples and take the variance of the full-observation $Q$ over the completed boards. The agent's belief over the masked features is the decoder averaged over its latent, $p(o_{N\setminus C} \mid o_C) = \E_{z}\, p_\theta(o_{N\setminus C} \mid z, o_C)$. Its entropy $\hat H(C)$, estimated by nested Monte Carlo, can be directly compared with $H$, unlike the entropy of $z$. We write $\hat\phi$ for the Shapley values of these estimated games.\looseness=-1
\section{Experiments}
\label{sec:exp}
To demonstrate the potential of a mask-aware belief network, we chose two environments with different characteristics. Tic-tac-toe ($n = 9$) is fully observable, and its inherent uncertainty comes from the opponent, which is either uniformly random or a perturbed minimax that plays a random move with a probability annealed from $100\%$ to $5\%$ during training. The agent plays either side, seeing its own marks as $+1$. Minesweeper $4\times4$ with 3 mines ($n = 16$) is partially observable, and its randomness comes only from mine positions. In both environments, exhaustive search of the game tree (expectimax) gives the optimal policy and its occupancy, on which all agents are scored. The exact games are computed for the optimal action under this occupancy, from each agent's own full-observation outputs, $Q$ for $\phi^Q$ and $\phi^{\mathrm{obs}}$ and $s$ for $\phi^{\mathrm{inh}}$, while $H$ depends on the occupancy alone.\looseness=-1

We compare with two baselines. DQN + surrogate (DQN + surr.) fits a surrogate to a standard DQN's $Q$ on masked inputs from its $\varepsilon$-greedy occupancy ($\varepsilon = 0.1$), and the mask-aware DQN (MA-DQN) is itself trained on masked inputs. Both also get our TD-trained variance head $s$ and a second-moment head $\widehat{Q^2}$, regressed on the squared full-observation $Q$. From their outputs on $o_C$, this gives $\hat v_{\mathrm{obs}}(C) = \max(\widehat{Q^2}(C) - \hat Q(C)^2, 0)$ and $\hat v_{\mathrm{inh}}(C) = \hat s(C)$. All MLPs have 2 hidden layers of width 256 with ReLU and the same training budget (EXPOSURE has about $4\times$ the parameters of the MA-DQN, but a width-512 MA-DQN of similar size does not close the gap). We enumerate all coalitions, so only EXPOSURE's Monte Carlo estimates add sampling error. Tic-tac-toe is scored on all $2190$ (random) and $1329$ (perturbed minimax) observations of the occupancy, Minesweeper on $300$ sampled from it. Our code is available at \url{https://github.com/kianaf/exposure-rl}.

\begin{figure}[t]
\centering
\includegraphics[width=0.9\textwidth]{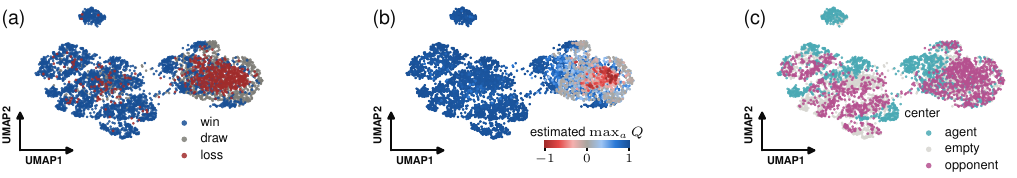}\\[2pt]
\includegraphics[width=0.9\textwidth]{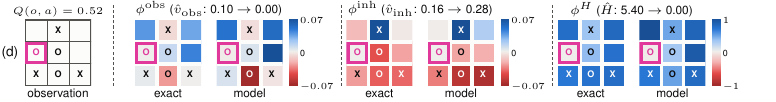}
\caption{Tic-tac-toe, random opponent. (a--c)~UMAP of the belief mean $\E[z \mid o_N]$ on all boards, colored by (a)~exact result, (b)~estimated $\max_a Q$ and (c)~who holds the key center cell. (d)~Exact and model Shapley values on one board (action in pink has no Shapley value). Headers give $\hat v(\emptyset) \rightarrow \hat v(N)$, and the model's Shapley values sum to $\hat v(\emptyset) - \hat v(N)$. Blue cells raise certainty, red ones lower it. Against the random opponent, the agent expects a win. Cells matter by the lines they lie on (center on four, a corner on three, an edge on two). In $\vobs$, free corners and the winning edge raise certainty, while the bottom O, an edge cell, lowers it, as it makes a draw likelier. In $\vinh$, the top X, a weak edge move by the opponent, raises certainty most. In $H$, each revealed cell narrows the belief.\looseness=-1}
\label{fig:ttt}
\end{figure}

\begin{table}[!b]
\centering\footnotesize
\setlength{\tabcolsep}{2.5pt}\renewcommand{\arraystretch}{0.88}
\begin{tabular*}{\textwidth}{@{\extracolsep{\fill}}llccccc@{}}
\toprule
 & & wins (\%) & $\phi^Q$: $\rho$ & $\phi^{\mathrm{obs}}$: $\rho$ & $\phi^{\mathrm{inh}}$: $\rho$ & $\phi^H$: $\rho$ \\
\midrule
\multirow{3}{*}{\begin{tabular}[c]{@{}l@{}}Tic-tac-toe\\random\end{tabular}} & EXPOSURE & $93.5 \pm 1.3$ & $\mathbf{0.67 \pm 0.01}$ & $\mathbf{0.64 \pm 0.05}$ & $\mathbf{0.69 \pm 0.01}$ & $\mathbf{0.76 \pm 0.00}$ \\
 & MA-DQN & $\mathbf{94.7 \pm 0.5}$ & $0.66 \pm 0.01$ & $0.43 \pm 0.01$ & $0.57 \pm 0.01$ & -- \\
 & DQN + surr. & $92.5 \pm 1.2$ & $0.63 \pm 0.01$ & $0.53 \pm 0.04$ & $0.49 \pm 0.02$ & -- \\
\midrule
\multirow{3}{*}{\begin{tabular}[c]{@{}l@{}}Tic-tac-toe\\perturbed\\minimax\end{tabular}} & EXPOSURE & $8.8 \pm 0.7$ & $\mathbf{0.67 \pm 0.01}$ & $\mathbf{0.42 \pm 0.02}$ & $\mathbf{0.85 \pm 0.00}$ & $\mathbf{0.56 \pm 0.01}$ \\
 & MA-DQN & $\mathbf{9.5 \pm 0.3}$ & $0.42 \pm 0.01$ & $0.23 \pm 0.01$ & $0.65 \pm 0.03$ & -- \\
 & DQN + surr. & $8.4 \pm 1.2$ & $0.49 \pm 0.02$ & $0.18 \pm 0.04$ & $0.59 \pm 0.03$ & -- \\
\midrule
\multirow{3}{*}{\begin{tabular}[c]{@{}l@{}}Minesweeper\\$4\times4$\end{tabular}} & EXPOSURE & $\mathbf{75.5 \pm 6.1}$ & $\mathbf{0.80 \pm 0.01}$ & $\mathbf{0.61 \pm 0.07}$ & $\mathbf{0.82 \pm 0.02}$ & $\mathbf{0.59 \pm 0.04}$ \\
 & MA-DQN & $49.5 \pm 11.9$ & $0.68 \pm 0.04$ & $0.15 \pm 0.06$ & $0.68 \pm 0.02$ & -- \\
 & DQN + surr. & $13.7 \pm 2.2$ & $0.66 \pm 0.01$ & $0.48 \pm 0.06$ & $0.64 \pm 0.01$ & -- \\
\bottomrule
\end{tabular*}

\caption{Mean per board Spearman correlation $\rho$ of each method's Shapley values and the exact ones, then mean $\pm$ SD over 3 seeds. Bold marks the largest mean, EXPOSURE's for every explanation, and only EXPOSURE gives $H$. Wins are greedy-policy win rates, and the optimal agent reaches $96.4$, $9.8$ and $87.9\%$.}
\label{tab:results}
\end{table}

\tabref{tab:results} gives the Spearman correlation $\rho$ between each method's Shapley values and the exact ones. EXPOSURE is best on the expected return and on the observational and inherent uncertainty in all three settings, with only a marginal lead on the former in tic-tac-toe with a random opponent. Only EXPOSURE gives $\hat\phi^H$, which agrees with the exact $\phi^H$ at $\rho \ge 0.56$. To check that the lead does not come from a better $Q$, we re-estimate the conditional means and variances of every method's $Q$ in the same way, over the boards that match $o_C$ among $10^4$ from the optimal policy's occupancy. The correlations then differ by at most $0.06$, consistent with coalition estimation contributing to the main gaps. This accuracy comes at no cost in performance, as EXPOSURE matches the baselines in tic-tac-toe and even wins more often in Minesweeper ($75.5\%$ versus $49.5\%$).\looseness=-1

In the trained tic-tac-toe agent, the latent is organized by result classes, i.e., win, loss, and draw against a perfect opponent, by expected return, and by who holds the center (\Figref{fig:ttt}a--c). A linear probe on the 16-dimensional belief mean recovers the result class with $69\%$ balanced accuracy versus $43\%$ for the same probe on the raw board, and the expected return with $R^2=0.57$ versus $0.20$. In \Figref{fig:ttt}d, the model's maps agree with the exact ones on the sign of each clearly colored cell and on the strongest cells of $\hat\phi^{\mathrm{obs}}$ and $\hat\phi^{\mathrm{inh}}$.
\section{Discussion and limitations}
\label{sec:conclusion}
EXPOSURE explains an agent's observational, inherent, and belief uncertainty with a mask-aware belief network that acts and explains, giving the most accurate explanations of every target it shares with the baselines. Three limitations remain. First, the decomposition does not cover epistemic uncertainty~\citep{wimmer2023quantifying}. Second, rank agreement with the exact Shapley values does not show calibration of learned beliefs and the variance head. Third, scaling to larger observation spaces needs amortized attribution such as FastSVERL~\citep{beechey2025fastsverl} and faithfulness tests without exact references. Comparing with a separately trained imputer~\citep{olsen2022vaeac} would show whether the belief must be learned jointly with the agent.\looseness=-1

\begin{footnotesize}
\bibliographystyle{unsrtnat}
\bibliography{refs}
\end{footnotesize}

\end{document}